\documentclass[11pt,a4paper]{article}
\usepackage[hyperref]{emnlp-ijcnlp-2019}
\usepackage{times}
\usepackage{latexsym}
\usepackage{subcaption}
\usepackage{graphicx}
\usepackage{multirow}

\usepackage{url}

\aclfinalcopy 

\newcommand{\cameraready}[1]{\textcolor{purple}{}}
\newcommand{\good}[1]{\underline{\textcolor{cyan}{#1}}}
\newcommand{\bad}[1]{\textit{\textcolor{red}{#1}}}

\title{How does Grammatical Gender Affect Noun Representations \\ in Gender-Marking Languages?}

\author{Hila Gonen$^1$ \space \space  Yova Kementchedjhieva$^2$ \space \space  Yoav Goldberg$^{1,3}$ \\
	$^1$Department of Computer Science,  Bar-Ilan University \\
	$^2$University of Copenhagen \\
	$^3$Allen Institute for Artificial Intelligence \\
	{\tt hilagnn@gmail.com,yova@di.ku.dk,yoav.goldberg@gmail.com} \\
}

\date{}

\begin{document}
\maketitle
\begin{abstract}

Many natural languages assign grammatical gender also to inanimate nouns in the language. In such languages, words that relate to the gender-marked nouns are inflected to agree with the noun's gender. We show that this affects the word representations of inanimate nouns, resulting in nouns with the same gender being closer to each other than nouns with different gender. 
While ``embedding debiasing'' methods fail to remove the effect, we demonstrate that a careful application of methods that neutralize grammatical gender signals from the words' context when training word embeddings is effective in removing it. Fixing the grammatical gender bias yields a positive effect on the quality of the resulting word embeddings, both in monolingual and cross-lingual settings. We note that successfully removing gender signals, while achievable, is not trivial to do and that a language-specific morphological analyzer, together with careful usage of it, are essential for achieving good results.

\end{abstract}

\section{Introduction}

Work on distributional word embeddings focuses almost exclusively on English, or on cross-lingual and language-agnostic techniques. However, languages are diverse and different languages exhibit different linguistic phenomena, which may interact with the English-centric embedding learning algorithms. In this work we look into one such phenomenon---grammatical gender---and examine its effect on the learned representation.

Many languages have rich grammatical systems, that often include a complex gender system as well \cite{C91}. Languages with grammatical gender assign and morphologically mark gender not only to animate nouns (which have biological sex, e.g. man, woman, mother, father), but also to inanimate nouns (e.g. dream, book). This grammatical gender assignment is mostly arbitrary: the same inanimate concept can have different gender in different languages. For example, a \emph{flower} is masculine in Italian (\emph{fiore}) and feminine in German (\emph{Blume}).

Languages often maintain an \emph{agreement system} in which certain words agree on different morphological features with other words they relate to. For example, English present-tense verbs are inflected to agree with their nominal subject on the \emph{number} feature. In other languages the agreement system is more elaborate, and in particular verbs, adjectives, determiners and other functions agree with nouns on many features, including gender \cite{C06}.\footnote{As the gender of nouns is fixed, the other elements are inflected to accommodate the agreement constraint. The nouns are said to \emph{assign gender} to the other words.} 

Such grammatical agreement affects the distributional environment of nouns, as nouns of different gender become surrounded by different word forms: feminine nouns co-occur more with the feminine forms of words, while masculine nouns with the masculine forms. For example, the Italian word \emph{viaggio} (``\emph{journey}''-masc) will co-occur with \emph{durato} (``\emph{last}''-masc) and \emph{lungo} (``\emph{long}''-masc), while the word \emph{gita} (``\emph{trip}''-fem) will co-occur with \emph{durata} (``\emph{last}''-fem) and \emph{lunga} (``\emph{long}''-fem).

Such changes in the distributional environment may bias the learned distributional representations of inanimate nouns. Indeed, we see that the majority of the top-10 nearest neighbors of the word \emph{gita} in Italian (``\emph{trip}''-fem) are feminine words. Also, we notice that the word \emph{viaggio} (``\emph{journey}''-masc) is not on the list, while in English, for comparison, we can find \emph{journey} in the top-10 nearest neighbors of \emph{trip}.

In this work, we are interested in investigating, demonstrating and quantifying this effect beyond the anecdotal level. We also explore methods for removing such unwanted biases.

We demonstrate that both in Italian and in German, the grammatical gender affects similarities between word representations (using words from SimLex-999 \cite{HRK15,LR15}): pairs of nouns with similar gender are closer to each other while pairs of nouns with different gender are farther apart.

After quantifying the effect, we explore several methods of reducing it. A popular choice would be to simply lemmatize all the words prior to feeding them to the embedding learning algorithm. However, full lemmatization can be destructive, in the sense that it will also remove morphological distinction that we may want to keep. We thus seek more surgical approaches. Interestingly, recent embedding debiasing approaches \cite{BCZ16} do not work well. We instead look for methods that attempt to neutralize the gender signals from the training data. We find that such methods are effective in reducing the effect, but are also language specific and tricky to get right: we rely on language specific morphological analyzers while carefully accounting for their peculiarities and adjusting our use for each language. We take this work as a reminder that (a) linguistic resources such as lexicons and morphological analyzers are still relevant and useful (cf. \cite{ZH17}); (b) languages are diverse and different languages require different treatments; and (c) small details may matter a lot. In particular, existing tools and resources, either learned or human curated, should not be trusted blindly, but be carefuly adapted for the problem.

Finally, we show that reducing the effect of grammatical agreement also has a positive effect on the quality of the resulting word representations, both in monolingual and cross-lingual settings.
We conclude that grammatical gender indeed has its imprints on the representations of inanimate nouns, and that this should be taken into account when working with gender-marking languages. Our code and debiased embeddings are available at \url{https://github.com/gonenhila/grammatical_gender}.

\section{Background and Related Work}

\paragraph{Word Embeddings}

Word embeddings have become an important component in many NLP models and are widely used for a vast range of downstream tasks. These models are based on the distributional hypothesis according to which words that occur in the same contexts tend to have similar meanings \cite{H54}. 
Indeed, they aim to create word representations that are derived from their shared contexts, where the context of a word is essentially the words in its proximity (be it according to linear order in the sentence or according to syntactic relations) \cite{MCC13,PSM14,LG14}. 

\paragraph{Gender Biases in Word Embeddings}

Social gender bias was demonstrated to be consistent and pervasive across different word embeddings \cite{CBN17}. Bolukbasi et al. \citeyearpar{BCZ16} show that using word embeddings for simple analogies surfaces many gender stereotypes. In addition, they define the gender bias of a word $w$ by its projection on the ``gender direction": $\overrightarrow{w} \cdot (\overrightarrow{he} - \overrightarrow{she})$, assuming all vectors are normalized. Positive bias stands for male-bias. For example, the bias of \textit{manager} is $0.06$, while the bias of \textit{nurse} is $-0.10$\footnote{in English word2vec embeddings \cite{MCC13} trained on Wikipedia.}.

Recently, some work has been done to reduce social gender bias in word embeddings, both as a post-processing step \cite{BCZ16} and as part of the training procedure \cite{ZZL18}. Bolukbasi et al. \citeyearpar{BCZ16} use a post-processing debiasing method. Given a word embedding matrix, they make changes to the word vectors in order to reduce the gender bias for all words that are not inherently gendered. They do that by zeroing the gender projection of each word on a predefined gender direction.\footnote{The gender direction is chosen to be the top principal component (PC) of ten gender pair difference vectors.}

In Zmigrod et al. \citeyearpar{ZMW19}, the authors mitigate social gender bias in gender marking languages using counterfactual data augmentation. Gender-marking languages add several interesting dimensions to the story: words relating to animate concepts such as ``nurse'' or ``cat'' may have both masculine and feminine versions; the distributional environment of a word contains many more explicit gender cues; and inanimate concepts are also assigned gender. All of these factors interact in complicated ways. In this work we focus on purely grammatical gender---the gender that is assigned to inanimate nouns---and its effect on their resulting representations.

\paragraph{Grammatical Gender Bias in Word Embeddings}

Grammatical gender is manifested in a similar way to social bias. For example, when projected on the Italian gender direction $\overrightarrow{lui} - \overrightarrow{lei}$ (Italian equivalents of ``he" and ``she"), the word ``secolo" (\emph{century}, masculine) has positive bias of 0.073, while the word ``zuppa" (\emph{soup}, feminine) has negative bias of -0.079.\footnote{in Italian word2vec embeddings \cite{MCC13} trained on Wikipedia.}

We attribute this behavior to grammatical agreement.
Since the context of different-gender nouns is expected to be very different because of the agreement of the surrounding words, and since the resulting representations are based on the context of the word, we expect grammatical gender to play a role in the representations---nouns with the same gender are expected to be closer together than nouns with different gender. For inanimate nouns, this behavior is undesired.

\paragraph{Word Embeddings and Morphology}

Word embeddings were shown to capture grammatical and morphological properties. Avraham and Goldberg \citeyearpar{AVG17} show that standard training of word embeddings in Hebrew captures also morphological properties and that using the lemmas when composing the representations helps to better capture semantic similarities. Similarly, Basirat and Tang \citeyearpar{BT18} show that typical grammatical features are captured by Swedish word embeddings. 

Cotterell et al. \citeyearpar{CSE16} treat the sparsity problem of morphologically rich languages in word embedding. They present a Gaussian graphical model to smooth representations of observed words and extrapolate representations for unseen words using morphological resources. With similar motivation, Vuli{\'c} et al. \citeyearpar{VMR17} use morphological constraints in English in order to pull inflectional forms of the same word closer together and push derivational antonyms farther apart. 
Finally, Salama et al. \citeyearpar{SYF18} enhance Arabic word embeddings by incorporating morphological annotations. 

\section{Grammatical Gender Affects Word Representations}
\label{effect}

As a first step, we aim to verify that the representation of inanimate nouns in gender-marking languages is indeed affected by their grammatical gender.
Since English does not have grammatical gender, a natural approach would be to use it as a reference when measuring this phenomenon.

\subsection{Inanimate Noun Pairs from SimLex-999}
We take the inanimate noun portion of the SimLex-999 dataset \cite{HRK15}, a gold standard resource for evaluating distributional semantic models. This dataset has an English version, and also German and Italian versions \cite{LR15}, and includes both similar and dissimilar word pairs, with human-assigned similarity judgments for each pair. This gives us 529 pairs of English words, along with high quality translations to Italian and German. We manually associate the Italian and German words with their grammatical gender.

\subsection{Differences in Similarities}
\label{diff_sim}

We divide the pairs in the gender-marking (GM) language (be it German or Italian) into two sets: (1) pairs of nouns that have the same gender in the GM language; (2) pairs of nouns that have different gender in the GM language. The respective English pairs are split in the same way, according to the gender of the nouns in the GM language. Thus, we end up with two sets of pairs in a GM language and their translations to English. Note that the English sets are different when used as a reference for German and Italian, since the split depends on the gender in the respective language.

For each set we compute the average of the cosine similarity of all word pairs within it. If gender plays a role in the representation of words, and indeed brings same-gender words closer together while keeping different-gender words farther apart, we expect to see a significant difference between the average similarity of the set of same-gender nouns and the set of different-gender nouns. As mentioned above, we compute these averages for English as a reference, where we expect a low difference between the two sets. Table~\ref{tab:effect_sim} shows the results for Italian and German, compared to English. Indeed, in both cases, the difference between the average of the two sets is much bigger. 

\begin{table}
	\begin{center}
	    \scalebox{0.85}{
		\begin{tabular} {l||lr|lr}
			& Italian & En & German & En \\ \hline \hline
			Same Gender & 0.442 & 0.424 & 0.491 & 0.446 \\\hline
			Different Gender & 0.385 & 0.415 & 0.415 & 0.403 \\\hline
			difference & 0.057 & 0.009 & 0.076 & 0.043 \\\hline
		\end{tabular}
		}
		\caption{Averages of similarities of pairs with same gender vs. different gender, along with the respective averages in English. The last row (difference) is the difference between the averages of the two sets.} 
		\label{tab:effect_sim}
	\end{center}
\end{table}

\subsection{Rank in Nearest Neighbor List}
\label{rank}

We take the same sets as before, and for each pair in them we compute the rank of the second word in the nearest neighbor list of the first word and vice versa. For example, for the pair ``parola" (\emph{word}) and ``dizionario" (\emph{dictionary}) in Italian, we compute the rank of ``dizionario" in the list of nearest neighbor of ``parola" and the rank of ``parola" in the list of nearest neighbors of ``dizionario".

We then compare the average ranking in each set, with English as the reference. If the gender affects the similarities between words, we expect same-gender pairs to have lower average than different-gender pairs (remember that the closest word is at the lowest rank: 1). Table~\ref{tab:effect_rank_fix} shows the results for Italian and German, compared to English. As expected, the average ranking of same-gender pairs is significantly lower than that of different-gender pairs, both for German and Italian, while the difference between the sets in English is much smaller.

\begin{table*}
    
	\begin{center}
	    \scalebox{0.7}{
		\begin{tabular} {l||c|c|c||c|c|c}
			&\multicolumn{3}{c||}{Italian} & \multicolumn{3}{|c}{German}  \\\hline
			& Same-gender & Diff-Gender & difference & Same-gender & Diff-Gender & difference \\ \hline \hline
			\multirow{3}{*}{7--10} & Og: 4884 & Og: 12947 & Og: 8063 & Og: 5925 & Og: 33604 & Og: 27679  \\
			& Db: 5523 & Db: 7312 & Db: 1789 & Db: 7653 & Db: 26071 & Db: 18418  \\
			& En: 6978 & En: 2467 & En: -4511 & En: 4517 & En: 8666 & En: 4149  \\ \hline
			\multirow{3}{*}{4--7} & Og: 10954 & Og: 15838 & Og: 4884 & Og: 19271 & Og: 27256 & Og: 7985  \\
			& Db: 12037 & Db: 12564 & Db: 527 & Db: 24845 & Db: 22970 & Db: -1875  \\
			& En: 15891 & En: 17782 & En: 1891 & En: 13282 & En: 17649 & En: 4367  \\ \hline
			\multirow{3}{*}{0--4} & Og: 23314 & Og: 35783 & Og: 12469 & Og: 50983 & Og: 85263 & Og: 34280  \\
			& Db: 26386 & Db: 28067 & Db: 1681 & Db: 60603 & Db: 79081 & Db: 18478  \\
			& En: 57278 & En: 53053 & En: -4225 & En: 41509 & En: 62929 & En: 21420  \\ \hline

		\end{tabular}
		}
		\caption{Averages of rankings of the words in same-gender pairs vs. different-gender pairs for Italian and German, along with their differences. \textbf{Og} stands for the original embeddings, \textbf{Db} for the debiased embeddings, and \textbf{En} for English. Each row presents the averages of pairs with the respective scores in SimLex-999 (0--4, 4--7, 7--10).} 
		\label{tab:effect_rank_fix}
	\end{center}
	
\end{table*}

\section{Debiasing Methods do not Work}

As mentioned above, grammatical gender bias shares some aspects with social gender bias. Keeping that in mind we first turn to use these existing methods of gender-debiasing in English word embeddings.

Bolukbasi's method \citeyearpar{BCZ16} requires sets of pairs that define the gender direction. For this we use their predefined pairs, since we target grammatical gender bias, which we have demonstrated to be similar to social gender bias. In addition, a predefined set of inherently-neutral words is also needed: these are the words that will be debiased by the algorithm.
As a first step, and in order to estimate the feasibility of using this method for reducing the grammatical gender bias, we use the set of the inanimate nouns from SimLex-999 as our set of inherently-neutral words.\footnote{If this method doesn't mitigate the bias we showed in the previous section, then using inherently-neutral words extracted from the vocabulary automatically cannot possibly work as well.}

The algorithm worked well in the sense that the bias of all inanimate nouns, when measured by their projection on the gender dimension, became zero. However, it also failed: the similarities between the inanimate nouns themselves hardly changed.
Table~\ref{tab:effect_sim_deb} depicts the average similarities in Italian before and after debiasing. 

\begin{table}
	\begin{center}
	    \scalebox{0.75}{
		\begin{tabular} {l||ccc|c}
			
			&\multicolumn{4}{c}{Italian}  \\\hline
			& Original & Debiased & English & Reduction \\ \hline \hline
			Same Gender & 0.442 & 0.439 & 0.424 & --  \\\hline
			Different Gender & 0.385 & 0.390 & 0.415 & -- \\\hline
			difference & 0.057 & 0.049 & 0.009 & \textbf{16.67\%} \\\hline

		\end{tabular}
        }
		\caption{Averages of similarities of pairs with same vs. different gender in Italian compared to the debiased version using Bolukbasi's \citeyearpar{BCZ16} method. The last row is the difference between the averages of the two sets. ``Reduction" stands for gap reduction after debiasing.} 
		\label{tab:effect_sim_deb}
	\end{center}
\end{table}

This suggests that the information about the gender is deeply embedded in the representation and is not easy to remove in a post-processing phase. Specifically, zeroing the projection of a word's vector on the gender direction is not enough in order to remove all gender information from the word's representation. The fact that similarities between words hardly change implies that the projection on the gender direction is not the only indication of gender. These results align with the findings discussed in Gonen and Goldberg \citeyearpar{GONEN19}. \cameraready{This result calls for a change during the training procedure. This is done by Zhao et al. \citeyearpar{ZZL18}. However, their method is also based on the projection of the words on the gender direction, which we showed to be an insufficient definition when dealing with gender bias.}

We conclude that focusing on the projection of vectors on the gender direction is not the right way to go, and we opt to removing gender inflections from the context before training. We describe this in detail in the next section.

\section{Removing Gender Inflection from the Context}
\label{removing}

As mentioned above, words in the surroundings of gender-marked nouns (e.g. articles, adjectives) are often inflected to agree with the gender of the noun they relate to. As we hypothesize that most of the effect shown in Section~\ref{effect} is caused by this gender agreement, we try several schemes that aim to remove gender signals from the context. 

A straight-forward approach would be to lemmatize all the words, which will remove all gender signals from the context of a word. However, this approach has two main downsides: 1) We would like to have a representation for all the words in the vocabulary, but changing also the target words will reduce the vocabulary size and result with missing words (we will no longer have different masculine and feminine forms for any word); 2) Lemmatization removes not only gender information, but also additional information (such as number and tense). While gender assignment is arguably arbitrary, and does not translate to an actual physical property of inanimate nouns in reality, other properties that agree with the noun, such as number, do hold in reality and signify actual properties of the target noun, which we prefer to preserve.

Thus, a better approach would be to neutralize gender signals from the context alone, keeping the target words intact. This way we do not change the resulting embedding vocabulary. This can be done using: 1) lemmatizing all the context words, where we lose additional information, as discussed above; 2) changing all the context words to the same gender, while keeping all other features of the words intact. Once the whole context is of the same gender, we essentially lose the gender information altogether as all nouns have similar context, regardless their gender.\footnote{Context nouns are also kept unchanged since nouns do not agree with other nouns in their context, both in Italian and in German. Notably, in German, we lose the noun-ness information when we lowercase the corpus (as all nouns in German begin with an uppercase letter).}

\subsection{The proposed approaches}

We experiment with both lemmatization of context words and gender change of context words.

\paragraph{Lemmatization of Context Words}

When training word2vec \cite{MCC13}, we use a morphological analyzer to identify the lemmas of words, and replace context words, but not target words, with their lemmas.

\paragraph{Gender Change of Context Words}

When training word2vec, we choose a gender (for example, masculine) and change all context words to that gender: each word that is identified as being of a different gender (in Italian: feminine, in German: feminine or neutral), is changed to its masculine form. This is also done using a morphological analyzer: when we identify a non-masculine analysis, we search for a masculine one that shares the same lemma and features.

In general, we found Italian to work better with gender change, and German to work better with lemmatization. We report full results in Section~\ref{results}.

\subsection{Challenges}

While conceptually simple, fully neutralizing gender information is more challenging than it initially appears, and requires careful attention to ``get right''. We describe some cases in which gender information can leak.

\paragraph{Human Curator Choices}
The morphological analyzer sometimes assigns different lemmas to an opposite-gender pair, as a result of human curater design choices.
For example, in Italian, ``delle" is the feminine of ``dei", but they are assigned the lemmas ``della" and ``del", respectively. 
Such cases leak gender signal in both cases of lemmatization and gender change: (1) When lemmatizing, each of the words gets a different lemma, manifesting the gender. (2) When changing the gender, the opposite-gender form of the word is not identified as these words do not share lemma, and the words stay unchanged.

This was very prominent in some high-frequency Italian words, and dealt with by fixing the analyzer: we identified all forms without a corresponding gendered-pair, manually aligned them, and assigned each pair a shared and unique lemma. This fix dramatically improved results when using either lemmatization or gender change.

\paragraph{Gender-Ambiguous Word Forms} Many word forms have several morphological analyses, resulting in different lemmas. Inspecting this ambiguity reveals two specific issues, in German and in Italian. First, many German words are ambiguous with respect to gender. For example, ``eine'' has a frequent feminine reading, but also a rare masculine one. When changing words to masculine, this word is identified as potentially masculine, and kept intact. The presence of the context word ``eine'' now leaks a feminine signal.\footnote{A possible solution would be to replace words with their lemmas whenever we identify both feminine and masculine analyses. This did not improve results in practice.}

Second, Italian has many cases of two words with a similar set of possible lemmas but with different gender. For example, ``usato" and ``usata" are masculine and feminine, respectively, and both have ``usare" and ``usato" as possible lemmas. 
If we select a consistent lemma for each word type, and end up selecting a different lemma for each of ``usato'' and ``usata'', we again leak signal regarding the original gender. 

One solution would be to use context-sensitive lemmatization, that chooses the correct analysis in context. However, doing this accurately is still an open problem. Our proposed solution is to randomly sample a lemma per word token. This improved lemmatization results in Italian by 25\%.

\paragraph{Multiple Opposite-gender Forms for a Word}

In some cases, a single word might have multiple forms in the opposite gender. For example, the Italian ``delle" is the feminine form of both ``dei" and ``degli", depending on the phonetic context. In this case, the former is much more common than the latter. A naive approach that chooses to convert ``delle" to ``degli" essentially keeps the feminine signal for these cases: every instance of ``delle" changes to ``delgli", which marks masculine nouns in much less common cases, while most masculine nouns are usually accompanied with the more common word ``dei".

Ideally, when changing the gender of a word, we want to change a word by another word with a similar frequency, otherwise, the gender signal will be manifested in the frequency mismatch, as in the example above.

We deal with this issue using the following heuristic: when changing to masculine form (or any other gender form), for each word we first find all its possible masculine forms. Then, we check the frequency of the original word in the corpus, and choose the option with the closest frequency to it. This indeed yields better results: when not addressing the frequency issue in Italian, we are able to reduce the effect only by 35.42\% (compared to 91.67\%, see Section~\ref{results} for more details).

\begin{figure*}
	\centering
	\begin{subfigure}{0.4\linewidth}
		\includegraphics[width=\linewidth]{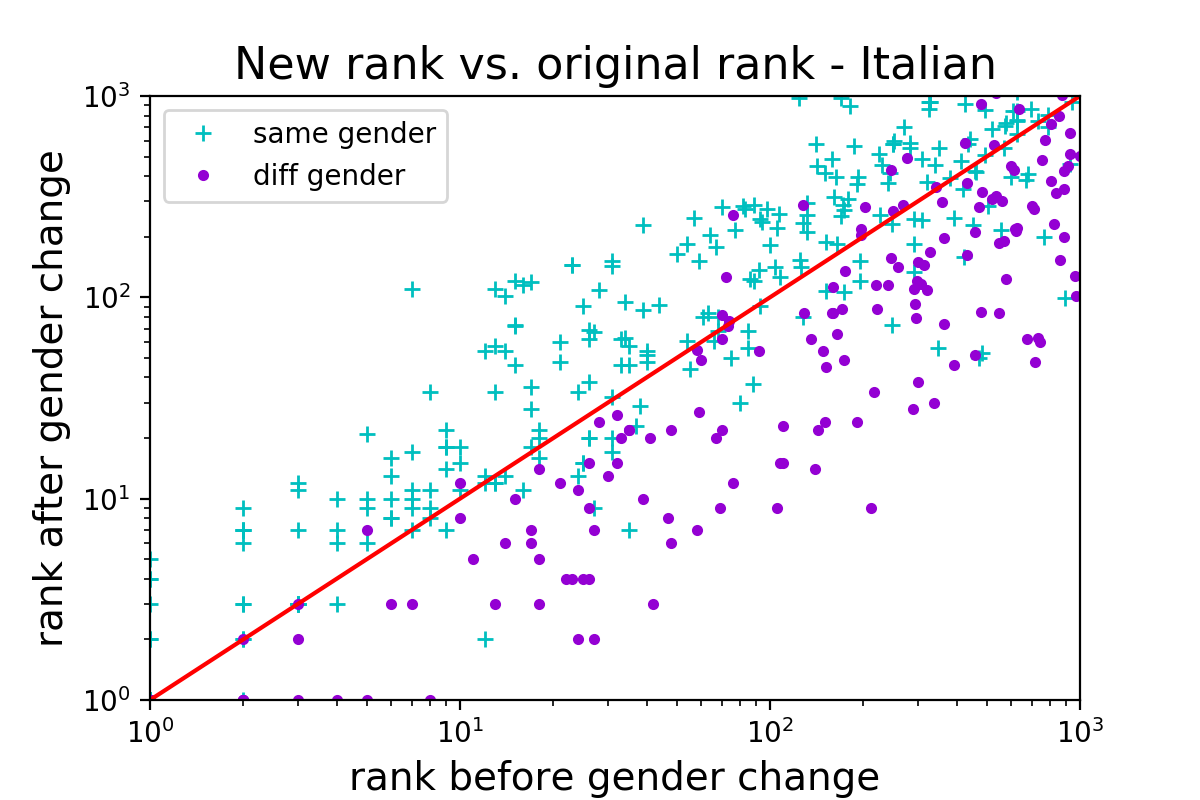}
		\caption{}
	\end{subfigure}
	\begin{subfigure}{0.4\linewidth}
		\includegraphics[width=\linewidth]{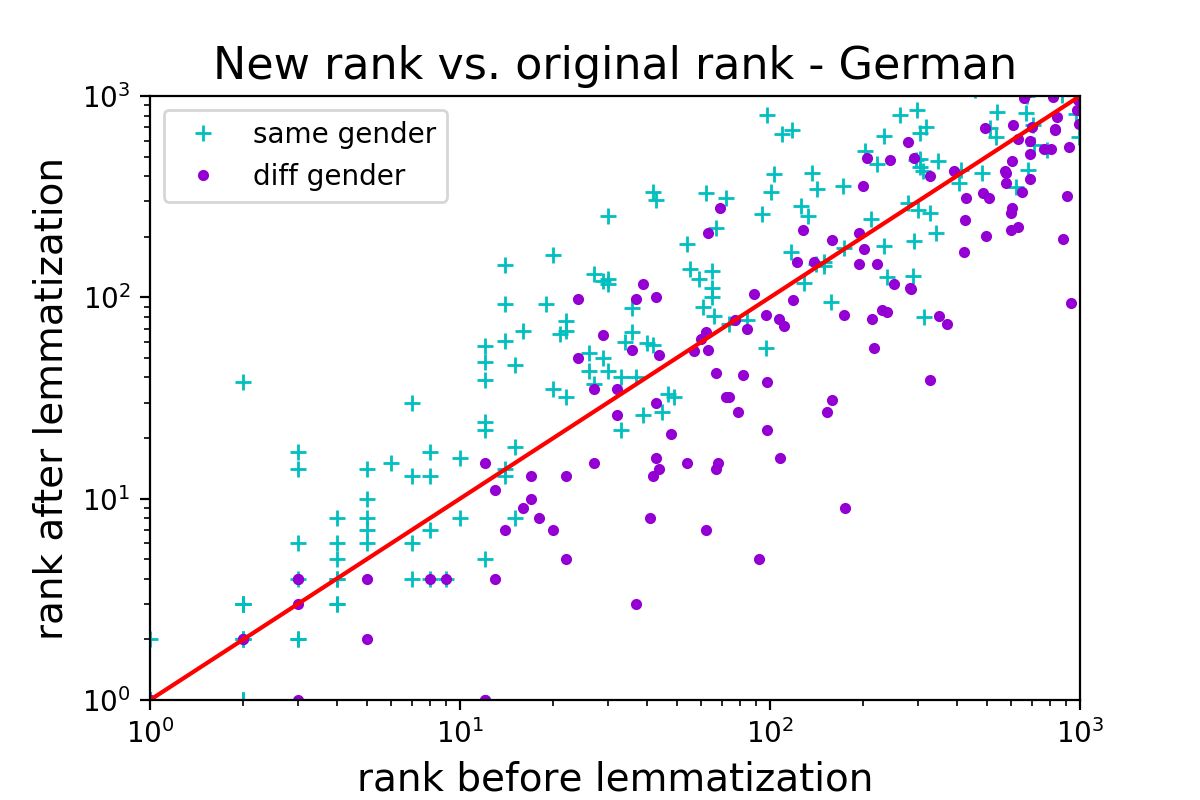}
		\caption{}
	\end{subfigure}
	\caption{The new rank of a word in the nearest neighbor list of its paired word. In cyan (+) -- pairs with the same gender, in purple ($\cdot$) -- pairs with different gender. Most words with same-gender are located above $y=x$ (were drifted apart), while most words with different-gender are located below it (got closer together).}
	\label{fig:graphs}
\end{figure*}

\section{Results}
\label{results}

We experimented with different schemes for each language, measuring their success at removing gender bias of inanimate nouns with respect to English.\footnote{We used state-of-the-art morphological analyzers for both languages. Full implementation details can be found in the appendix.}

For German, we found lemmatization to work better than gender change. In Italian gender change got better results. Specifically, changing to feminine got much better results than changing to masculine, probably due to less ambiguity when changing to feminine in some very common articles (see full manual mapping in the Appendix), resulting in fewer cases in which the gender signal leaks through the frequencies of the changed words, as explained above. In addition, the manual fixes to the lemmatizer were crucial to get satisfying results for both methods. 

While some of these findings depend on the specific morphological analyzer in use, the challenges and issues we demonstrate are relevant in any case.

\subsection{Reduction in Gender Bias}

\paragraph{Differences in Similarities}
We repeat the experiment in Section~\ref{diff_sim}---computing the average of pair similarities in each of the sets defined in Section~\ref{effect}, this time with the embeddings trained after removing gender signal from the context (debiasing). Table~\ref{tab:effect_sim_fix} shows the results for Italian and German, compared to English, both for the original and the debiased embeddings (for each language we show the results of the best performing debiased embeddings). As expected, in both languages, the difference between the average of the two sets with the debiased embeddings is much lower. In Italian, we get a reduction of 91.67\% of the gap with respect to English. In German, we get a reduction of 100\%. Note that for both languages, the main change is in the set of different-gender pairs, and not in the same-gender pairs. This makes sense as same-gender words have similar contexts both before and after our intervention, but different-gender words have different contexts before, but much more similar contexts after.

\begin{table*}
	\begin{center}
	    \scalebox{0.75}{
		\begin{tabular} {c||ccc|c||ccc|c}
			
			&\multicolumn{4}{c||}{Italian} & \multicolumn{4}{|c}{German}  \\\hline
			& Original & Debiased & English & Reduction & Original & Debiased & English & Reduction\\ \hline \hline
			Same Gender & 0.442 & 0.434 & 0.424 & -- & 0.491 & 0.478 & 0.446 & -- \\\hline
			Different Gender & 0.385 & 0.421 & 0.415 & -- & 0.415 & 0.435 & 0.403 & -- \\\hline
			difference & 0.057 & 0.013 & 0.009 & \textbf{91.67\%} & 0.076 & 0.043 & 0.043 & \textbf{100\%}  \\\hline

		\end{tabular}
        }
		\caption{Averages of similarities of pairs with same vs. different gender in Italian and German compared to English. The last row is the difference between the averages of the two sets. ``Reduction" stands for gap reduction when removing gender signals from the context.} 
		\label{tab:effect_sim_fix}
	\end{center}
\end{table*}

For comparison, in Italian we got 12.50\% reduction when using the lemmatization scheme, and 68.75\% reduction when lemmatizing with the addition of the manual mapping. For German, the best result using gender change was a reduction of 48.48\%, achieved by changing to neutral.

\paragraph{Rank in Nearest Neighbor List}
We repeat the experiment shown in Section~\ref{rank}---for each pair we compute the rank of the second word in the nearest neighbor list of the first word and vice versa. Then we compare the average ranking in each of the defined sets. Table~\ref{tab:effect_rank_fix} shows the results for Italian and German, both for the original and the debiased embeddings. As we expect, the difference between the average ranking of the two sets drops significantly for both languages.

In order to get a better picture of how the rankings of the different words change as a result of the gender signal removal, we take all pairs (and the inverted pairs). For each pair we plot the new rank of the second word in the nearest neighbors list of the first word as a function of its original rank before debiasing. Points above $y=x$ are of words that got a higher rank (lower in the list, farther from the first word), while points below this line are of words that got a lower rank (higher in the list, closer to the first word). Figure~\ref{fig:graphs} shows these plots for Italian and German. As expected, most words of same-gender pairs are located above the line (were drifted apart), while most words of different-gender pairs are located below the line (got closer together).

\subsection{Improvement in Word Similarities}

\paragraph{Qualitative Evaluation}
As a qualitative evaluation, we take several words for SimLex-999 and look at their top-10 nearest neighbor lists, before and after applying our method. In Table~\ref{tab:top10} we show the top-10 lists for the words \emph{vaso} (``\emph{jar}"-masculine) in Italian, and \emph{welt} (``\emph{world}"-feminine) in German. It is evident that the words that are added to the list, are better correlated with the target word than those that are removed. Two additional words appear in the Appendix.

\begin{table*}
    
	\begin{center}
	    \scalebox{0.8}{
		\begin{tabular} {  l|l||l|l }
			\multicolumn{2}{c||}{ Italian} & \multicolumn{2}{|c}{German}  \\\hline
			
			\multicolumn{2}{c||}{vaso \small(jar-masculine)} & \multicolumn{2}{c}{welt \small(world-feminine)}  \\\hline
            Orig & Debias & Orig & Debias \\ \hline
            
            coccio & vasi & welt" & europas \\
            recipiente & \good{ciotola} & europas & welt" \\
            otre & \good{bacinella} \small(basin) & scheibenwelt & scheibenwelt \\
            cinerario & recipiente & h\"asslichsten & \good{universum} \small(universe) \\
            vasetto & coccio & \bad{erde} \small(earth)& \good{menschheitsgeschichte} \small(human history)\\
            \bad{bacile} \small(basin) & cinerario & \bad{weltgeschichte} \small(world history) & h\"asslichsten \\
            \bad{kantharos} & otre & \bad{kl\"ugste} \small(wisest) & \good{menschheit} \small(mankind) \\
            vasi & vasetto & \bad{kl\"ugsten} \small(wisest) & schwarzafrikas \\
            \bad{vassoio} \small(tray) & \good{brocca} \small(pitcher) & schwarzafrikas & \good{parallelwelten} \small(parallel worlds) \\
            \bad{coperchio} \small(cover) & \good{scodella} \small(bowl) & \bad{lustigsten} \small(funniest) & \good{ulldart} \\

		\end{tabular}
		}
		\caption{Examples of top-10 nearest neighbor lists for words in Italian and in German, before and after debiasing. In red (italic) are words that were removed from the list, and in blue (underlined) are words that were added to it. Translations to English (Google Translate) for the changed words are in parenthesis, when different from source.}
		\label{tab:top10}
	\end{center}
	
\end{table*}

\paragraph{Evaluation on Simlex and WordSim-353}
We evaluate the quality of the grammatical-gender-neutralized embeddings using two datasets for each language: SimLex-999 \cite{HRK15,LR15} and WordSim-353 \cite{FGM02,LR15}. Table~\ref{tab:quality} shows the results for Italian and German for both datasets, compared to the original embeddings. In both cases, the new embeddings perform better than the original ones.

\begin{table}

	\begin{center}
	\scalebox{0.8}{
		\begin{tabular} {l||c|c||c|c}
			&\multicolumn{2}{c||}{ Italian} & \multicolumn{2}{|c}{German}  \\\hline
			& Orig & Debias & Orig & Debias \\ \hline \hline
			SimLex & 0.280 & \textbf{0.288} & 0.343 & \textbf{0.356}  \\\hline
			WordSim & 0.548 & \textbf{0.577} & 0.547 & \textbf{0.553}  \\\hline

		\end{tabular}
    }
		\caption{Results on SimLex-999 and WordSim-353, in Italian and German, before and after debiasing.} 
		\label{tab:quality}
	\end{center}

\end{table}

\paragraph{Cross-lingual Word Embeddings}

Studies in language and cognition suggest that humans share a common semantic space, regardless of their native language \cite{YSS16}. To the extent that embeddings capture the semantics of words, we can thus expect embedding spaces to have a similar structure across languages. Youn's statement concerns concepts and not words, however, and concepts can surface in many different forms in language, which interferes with how well embedding spaces align across languages \cite{SRV18}. Thus, we expect grammatical gender to have a negative impact on alignability.

We explore this matter through the task of cross-lingual embedding alignment, wherein a cross-lingual embedding space is learned through an alignment of independently pre-trained monolingual embeddings for a directed pair of languages. The quality of cross-lingual embeddings learned this way can be evaluated intrinsically on the task of bilingual dictionary induction (BDI). BDI queries the cross-lingual embedding space with a seed of words in one language, retrieves their counterparts among the words in the other language\footnote{This is done by minimizing a distance metric, most commonly, CSLS \cite{CLR18}.} and evaluates the precision of the produced translations against a set of gold standard targets. We carry out experiments using the supervised variant of the MUSE embedding alignment system \cite{CLR18} and report results on the inanimate portion of SimLex-999. We train a cross-lingual embedding alignment between English and either German or Italian, using the original and the debiased embeddings for these two languages. The results reported in Table~\ref{tab:cross} show that precision on BDI indeed increases as a result of the reduced effect of grammatical gender on the embeddings for German and Italian, i.e. that the embeddings spaces can be aligned better with the debiased embeddings.

\begin{table}

	\begin{center}
	    \scalebox{0.8}{
		\begin{tabular} {l||c|c||c|c}
			&\multicolumn{2}{c||}{ Italian} & \multicolumn{2}{|c}{German}  \\\hline
			& $\rightarrow$ En & En $\rightarrow$ & $\rightarrow$ En & En $\rightarrow$ \\ \hline \hline
			Orig  & 58.73 & 59.68 & 47.58 & 50.48 \\\hline
			Debias & \textbf{60.03} & \textbf{60.96} & \textbf{47.89} & \textbf{51.76} \\\hline

		\end{tabular}
        }
		\caption{Cross-lingual embedding alignment in Italian and in German, before and after debiasing.} 
		\label{tab:cross}
	\end{center}

\end{table}

\section{Conclusion}

We show that grammatical gender impacts word embeddings of inanimate nouns, both in Italian and in German, causing the similarities between words to change according to having same or different gender: the representations of same-gender words are closer together than representations of different-gender words.

We show that this effect can be almost completely removed when neutralizing gender signals in the context during training of the word embeddings. While most works in our field nowadays try to be language-independent, this is not always the right way to go: successfully removing those gender signals is not trivial to do and a language-specific morphological analyzer, together with careful usage of it, are essential for achieving good results.\footnote{Indeed, before implementing the specific fixes described in Section \ref{removing}, the reduction compared to English when naively changing to masculine was substantially smaller, 35.42\% reduction compared to 91.67\% in Italian, and 12.12\% compared to 100.00\% (with lemmatization) in German.}

In addition, this work serves as a reminder that languages other than English have different properties that are rarely dealt with when processing English. These aspects should be taken into account when dealing with morphologically reach languages, as not all models and algorithms for English can transfer directly to other languages.

\section{Acknowledgements}

The work was supported by the Israeli Science Foundation (grant number 1555/15) and by and the European Research Council (ERC Starting Grant iExtract 802774). We thank Valentina Pyatkin for the help with the Italian manual mapping.

\bibliography{bib_full}
\bibliographystyle{acl_natbib}

\clearpage

\appendix

\section{Implementation Details}

\paragraph{Morphological Analyzers}
For Italian, we use Morph-it!,\footnote{\url{http://tools.sslmit.unibo.it/doku.php?id=resources:morph-it}} a lexicon of inflected forms with their lemma and morphological features.
For German, we use DEMorphy,\footnote{\url{https://github.com/DuyguA/DEMorphy}} which, given a word, provides its full morphological analysis (or several, when applicable) \cite{A18}.\footnote{Since the analysis is fine-grained, when searching for a different-gender word form, we do not require full match, but restrict ourselves only to the following categories: CATEGORY, NUMERUS, PERSON, PTB\_TAG and TENSE.}

\paragraph{Training Word Embeddings}
We train 300d word embeddings with window size 4, on January 2018 wikipedia dump\footnote{https://dumps.wikimedia.org/} for all three languages. After tokenization we get 2.2B (En), 463M (It) and 815M (De) tokens. We discard words that do not appear at least 50 times, and are left with vocabulary sizes of 360,386 (En), 161,144 (It) and 361,944 (De).
We train using word2vecf \cite{LG14}, which allows to change context words without affecting target words.

\section{Manual Mapping for Italian}

Tables~\ref{gender} and ~\ref{lemma} contain the manual mappings we used for Italian (see next page).

\section{Qualitative Evaluation}

As a qualitative evaluation, we take several words for SimLex-999 and look at their top-10 nearest neighbor lists, before and after applying our method. In Table~\ref{tab:top10_sec} we show the top-10 lists for the words \emph{palla} (ball-feminine) in Italian, and \emph{diamant} (diamond-masculine) in German. It is evident that the words that are added to the list are better correlated with the target word than those that are removed (see next page).

\clearpage

\begin{table}[!h]
	
	\begin{center}
		\scalebox{0.7}{
			\begin{tabular} { l|l }
				
				word & opposite-gender form \\ \hline
				alla & [al, allo] \\
				alle & [agli, ai] \\
				colei & [colui] \\
				costei & [costui] \\
				esse & [essi] \\
				dalla & [dallo, dal] \\
				della & [dello, del] \\
				delle & [dei, degli] \\
				essa & [esso, egli] \\
				impostala & [impostalo] \\
				impostale & [impostagli] \\ 
				la & [lo, gli, il] \\
				lei & [lui] \\
				nella & [nel, nello]\\
				nelle & [nei, negli]\\
				ognuna & [ognuno]\\
				provocatele & [provocategli]\\
				qualcuna & [qualcuno]\\
				dalle & [dagli, dai]\\
				sulla & [sullo, sul]\\
				sulle & [sui, sugli]\\
				una & [un, uno]\\
				un' & [un]\\
				le & [lo, i, li]\\
				riformatorie & [riformatori]\\
				scatenatasi & [scatenatosi]\\
				scatenatesi & [scatenato]\\
				tutorie & [tutorio]\\
				al & [alla]\\
				agli & [alle]\\
				ciascuno & [ciascuna]\\
				colui & [colei]\\
				costui & [costei]\\
				dagli & [dalle]\\
				dallo & [dalla]\\
				dei & [delle]\\
				dello & [della]\\
				essi & [esse]\\
				esso & [essa]\\
				i & [le]\\
				impostagli & [impostale]\\
				impostalo & [impostala]\\
				li & [le]\\
				lo & [la, le]\\
				lui & [lei]\\
				nei & [nelle]\\
				nel & [nella]\\
				ognuno & [ognuna]\\
				provocategli & [provocatele]\\
				qualcuno & [qualcuna]\\
				sui & [sulle]\\
				sullo & [sulla]\\
				ultro & [ultra]\\
				ai & [alle]\\
				allo & [alla]\\
				dai & [dalle]\\
				dal & [dalla]\\
				degli & [delle]\\
				del & [della]\\
				egli & [essa]\\
				gli & [la]\\
				il & [la]\\
				negli & [nelle]\\
				sugli & [sulle]\\
				sul & [sulla]\\
				nello & [nella]\\
				un & [una, un']\\
				uno & [una]

			\end{tabular}
		}
		\caption{word: opposite-gender-form mapping for Italian.} 
		\label{gender}
	\end{center}
\end{table}

\begin{table}
	
	\begin{center}
		\scalebox{0.8}{
			\begin{tabular} { l|l }
				
				word & lemma \\ \hline
				un', un, una, uno & lemma1\\
				la, lo, gli, il & lemma2\\
				le, i, li & lemma3\\
				alla, al, allo & lemma4\\
				alle, agli, ai & lemma5\\
				colei, colui& lemma6\\
				costei, costui& lemma7\\
				esse, essi& lemma8\\
				dalla, dallo, dal& lemma9\\
				della, dello, del& lemma10\\
				delle, dei, degli& lemma11\\
				essa, esso, egli& lemma12\\
				impostala, impostalo& lemma13\\
				impostale, impostagli& lemma14\\
				lei, lui& lemma15\\
				nella, nel, nello& lemma16\\
				nelle, nei, negli& lemma17\\
				ognuna, ognuno& lemma18\\
				provocatele, provocategli& lemma19\\
				ualcuna, qualcuno& lemma20\\
				dalle, dagli, dai& lemma21\\
				sulla, sullo, sul& lemma22 \\
				sulle, sui, sugli& lemma23\\
				riformatorie, riformatori& lemma24\\
				scatenatasi, scatenatosi& lemma25\\
				scatenatesi, scatenato& lemma26\\
				tutorie, tutorio& lemma27\\ 
				ciascuno, ciascuna& lemma28\\
				ultro, ultra& lemma29

			\end{tabular}
		}
		\caption{word:lemma mapping for Italian.} 
		\label{lemma}
	\end{center}
\end{table}

\begin{table*}
	
	\begin{center}
		\scalebox{0.9}{
			\begin{tabular} {  l|l||l|l }
				\multicolumn{2}{c||}{ Italian} & \multicolumn{2}{|c}{German}  \\\hline
				
				\multicolumn{2}{c||}{palla \small(ball-feminine)} & \multicolumn{2}{c}{diamant \small(diamond-masculine)} \\\hline
				Orig & Debias & Orig & Debias \\ \hline
				
				pallina & pallina & smaragd & smaragd \\
				\bad{biglia} \small(ball) & pallone & topas & korund \\
				racchetta & \good{canestro} \small(basket) & ultramarin & topas \\
				pallone & \good{rimbalzo} \small(rebound) & salmiak & ultramarin \\
				\bad{stecca} \small(splint)& racchetta & \bad{gr\"unspan} \small(verdigris) & vitriol\\
				\bad{bilia} \small(marble)& \good{calciato} \small(kicked) & vitriol & \good{saphir} \small(sapphire) \\
				\bad{dapples} & schivata & korund & salmiak \\
				\bad{bandierina} \small(pennant)& \good{guantone} \small(mitt) & \bad{titan} \small(titanium) & \good{perle} \small(pearl) \\
				schivata & battitore & aquamarin & aquamarin \\
				battitore & \good{calciando} \small(kicking) & \bad{bornitrid} \small(boron nitride) & \good{unedlen} \small(base) \\
				
			\end{tabular}
		}
		\caption{Examples of top-10 nearest neighbor lists for words in Italian and in German, before and after debiasing. In red (italic) are words that were removed from the list, and in blue (underlined) are words that were added to it.
			Translations to English (Google Translate) for the changed words are in parenthesis, when different from source.} 
		\label{tab:top10_sec}
	\end{center}
	
\end{table*}

\end{document}